\pdfoutput=1

\documentclass[11pt]{article}

\usepackage[]{EMNLP2022}

\usepackage{times}
\usepackage{latexsym}
\usepackage{booktabs}
\usepackage{graphicx}
\usepackage{amsmath}
\usepackage{float}
\usepackage{xcolor}
\usepackage{adjustbox}
\usepackage{longtable}

\usepackage{placeins} 
\usepackage{microtype} 
\usepackage{enumitem} 
\usepackage{amsmath, bm} 
\usepackage{comment} 

\usepackage[T1]{fontenc}

\usepackage[utf8]{inputenc}

\usepackage{microtype}

\usepackage{inconsolata}

\title{Named Entity Linking with Entity Representation by Multiple Embeddings}

\author{Oleg Vasilyev, Alex Dauenhauer, Vedant Dharnidharka, John Bohannon \\
  Primer Technologies Inc. \\
  San Francisco, California \\
  \texttt{{oleg,alex.dauenhauer,john}@primer.ai}\\}

\begin{document}
\maketitle
\begin{abstract}
We propose a simple and practical method for named entity linking (NEL), based on entity representation by multiple embeddings. To explore this method, and to review its dependency on parameters, we measure its performance on Namesakes, a highly challenging dataset of ambiguously named entities. Our observations suggest that the minimal number of mentions required to create a knowledge base (KB) entity is very important for NEL performance. The number of embeddings is less important and can be kept small, within as few as 10 or less. We show that our representations of KB entities can be adjusted using only KB data, and the adjustment can improve NEL performance. We also compare NEL performance of embeddings obtained from tuning language model on diverse news texts as opposed to tuning on more uniform texts from public datasets XSum, CNN / Daily Mail. We found that tuning on diverse news provides better embeddings.
\end{abstract}

\section{Introduction}
Named entity linking (NEL) is a task of linking a mention of an entity in a text to the correct reference entity in the knowledge base (KB) \cite{rao2012entity, yang2015smart, sorokin-gurevych-2018-mixing, kolitsas2018endtoend, Logeswaran2019ZeroShotEL, Wu2020ScalableZE, Li2020Efficient, Sevgili2021Neural}. Here we consider NEL in a specific setting, with the intention to present our NEL method, and to probe the difficulty of dealing with namesakes:
\begin{enumerate}[topsep=0pt,itemsep=-1ex,partopsep=1ex,parsep=1ex]
    \item The mention of interest is assumed to be located in the text, i.e. the named entity recognition task is done.
    \item Only the local context surrounding the mention of interest is used for the linking, no other mentions in the text are used.
    \item KB is fixed and built on reliable data.
    \item Both KB and the pool of mentions are mostly composed of namesakes.
    \item A mention may have the corresponding entity in KB, or may not. We call the former mention \textit{familiar}, and the latter \textit{stranger}.
\end{enumerate}
The point 1 means that the named entity recognition task is assumed to be done, leaving us NEL in a narrow sense \cite{rao2012entity, Wu2020ScalableZE, Logeswaran2019ZeroShotEL}. The point 2 makes the problem better defined. Using other mentions of the same entity or related entities can help NEL, but our focus is on imitating the more difficult cases of lone mentions (with no related mentions in the vicinity in the text). For using related named entities see for example \cite{zaporojets-etal-2022-towards}. 

The point 3 leaves out the question of growing or improving KB by the encountered mentions (familiar or stranger). KB built on reliable data allows to isolate the effect of KB \textit{pollution} by intentionally adding wrong data, as we will do in this paper. The point 4 makes it easier to reveal NEL errors and to observe dependencies of our method on parameters.

Both the points 3 and 4 are satisfied by choosing recent dataset \textbf{Namesakes} \cite{namesakes2021, Vasilyev2021Namesakes}, as a dataset with human labeled ambiguously named entities. 
Another recent dataset  - Ambiguous Entity Retrieval (AmbER) \cite{Chen2021Evaluating} - does include subsets of identically named entities (for the purpose of fact checking, slot filling, and question-answering tasks), but it is automatically generated. Most existing NEL-related datasets do not focus on highly ambiguous names \cite{ratinov-etal-2011-local, hoffart-etal-2011-robust, Ferragina2012FastAA, Ji2017OverviewOT, guo2018}. 

In this paper we focus on presenting our NEL method. We test it on KB entities and mentions taken from Namesakes: the challenging dataset is helpful in revealing the behavior of our NEL method and its dependencies on parameters. Our contribution:
\begin{enumerate}[topsep=0pt,itemsep=-1ex,partopsep=1ex,parsep=1ex]
    \item We introduce a simple and practical representation of entity in KB, and explore NEL to such representations on example of highly ambiguous mentions from Namesakes.
    \item We suggest an adjustment of KB based on its entities, and show that it helps in reducing NEL errors.
\end{enumerate}
In Section \ref{sec:NEL} we introduce our entity representation and NEL for KB with such representations. In Section \ref{sec:Evaluation_on_Namesakes} we explain how we use Namesakes dataset in our NEL evaluation experiments. In Section \ref{sec:Experiments} we present the experiments and results.

\section{Named Entity as a Set of Embeddings}
\label{sec:NEL}
\subsection{Knowledge Base Entity}
\label{sec:KB_Entity}
A named entity can be described in very different contexts \cite{ma-etal-2021-muver, fitzgerald-etal-2021-moleman}. The same person can be a scientist and a dissident, the same location can be described by its nature and by its social events and so on. This is the motivation to represent an entity by multiple embeddings - at least if each embedding is created from a mention in specific context. 

Our representation of KB entity $E_a$ is composed of multiple embeddings, and consists of:
\begin{enumerate}[topsep=0pt,itemsep=-1ex,partopsep=1ex,parsep=1ex]
    \item Normalized embeddings $e_i$, their norms $|e_i|$ and assigned thresholds $t_i$, initially set $t_i=-1$. The number of the embeddings is restricted by agglomerative clustering (Appendix \ref{sec:Appendix_Settings}), $i<=N_E$.
    \item Entity threshold $T$.
    \item Entity surface names $s_k$.
    \item References to similar entities $E_b$.
\end{enumerate}
In this and the next subsections we explain the details of this representation, and our NEL procedure that uses it.

A fundamental element used in building an entity is an embedding of a mention of this entity in some context. We tuned a pretrained BERT \cite{Devlin2018BERT} language model ’bert-base-uncased', accessed via the transformers library (\citep{Wolf2020Transformers}), on generic random news, with named entities located in the texts. The only goal of tuning is to enhance LM performance on the mentions of named entities, without changing LM goal or making LM specialized on any particular set of named entities. The tuning and inference are as following:
\begin{enumerate}[topsep=0pt,itemsep=-1ex,partopsep=1ex,parsep=1ex]
    \item At tuning only the named entities - all the located mentions within the input window in the text - serve as the labels for prediction. In input the mentions are being either left unchanged (probability 0.5) or replaced by another random mention from the same text.
    \item At inference the text is kept as it is, and the model is run only once on each input-size chunk of the text. The embeddings are picked up from the first token of the entity surface form - for each named entity that happened to occur in the input-size chunk.
\end{enumerate}
Through all the paper, except Section \ref{sec:ModelXSumCNNDM}, we use the model tuned on random generic news. In Section \ref{sec:ModelXSumCNNDM} we use the model tuned on texts of more uniform style: the texts from XSum \cite{narayan-etal-2018-dont} and CNN / Daily Mail \cite{Hermann:2015:Teaching, cnndailymail} datasets. This allowed us to observe effect of using embeddings from a model exposed to a lesser variety of styles.
For more details of the tuning see Appendix \ref{sec:Appendix_TuningLM}.

There can be many mentions available for building a KB entity, even if only reliable verified mentions are used. If the number of the embeddings obtained from the mentions is higher than $N_E$, the embeddings are clustered, and only $N_E$ 'central' embeddings (closest to the centers of the clusters) are stored. We use agglomerative clustering (Appendix \ref{sec:Appendix_Settings}). The surface names $s_k$ of all the mentions used for creating an entity are stored in the entity.

\subsection{Linking a Mention to KB}
\label{sec:Linking_Mention_to_KB}

In linking a mention to KB we use only the normalized embedding $e$ of the mention. We define similarity $S$ of the mention to an entity by the scalar products with its embeddings $e_i$:
\begin{equation}
\label{eq:Similarity}
  S = \max_i[(e*e_i) / max(T, t_i)]
\end{equation} 
Here all the thresholds $t_i$ are set to $-1$, and are irrelevant unless adjusted as described in Section \ref{sec:KB_Adjustment}; the entity's threshold $T$ is defined further below.

The KB entity with the highest similarity $S$ is the candidate for linking the mention to. We set a \textit{linking threshold} $T_L$: The mention is linked to the candidate-entity only if 
\begin{equation}
\label{eq:Threshold_Global}
  S >= T_L
\end{equation}
Otherwise the mention is left \textit{unlinked} (unassociated with any KB entity). It is natural to assume $T_L=1$, but our results will show that we had to lower it.

The entity's threshold $T$ is defined from the assumption that any embedding of the entity would have to successfully link to the entity (with $T_L=1$):
\begin{equation}
  T = \min_j[\max_{i \neq j}[(e_j*e_i)]]
\end{equation} 
This definition makes sense only if there are at least two embeddings in the entity, hence we create a KB entity only if there are at least two mentions available (our observations in Section \ref{sec:Experiments} suggest a more strict requirement).

When linking a mention to KB we find the similarities not to all KB entities, but only to the entities that have a surface name at least somewhat similar to the mention. For this purpose, KB stores a map of all words from all the surface names to the entities that have any such word in their surface names:
\begin{equation}
  map: w \rightarrow \{E_a | w \; \text{in} \; s_k \; \text{in} \; E_a\}
\end{equation} 
When linking a mention to KB, all KB entities that are mapped from at least one word of the mention's surface name are considered as the candidates for linking. The similarities of the mention's embedding to these candidates are calculated then by Eq.\ref{eq:Similarity}; the mention is linked to the candidate with the strongest similarity if exceeds the threshold Eq.\ref{eq:Threshold_Global}.
Generally, KB entities with similar (by some measure) embeddings can be considered in selecting the candidates, but we focus here on the namesakes.

\subsection{Knowledge Base Adjustment}
\label{sec:KB_Adjustment} 
Can we improve KB right after it is created, even without using any knowledge about the texts and mentions on which NEL will be used or evaluated? We suggest adjusting the thresholds $t_i$ in each entity $E_a$ by considering the relation of $E_a$ with its similar entities $E_b$.

Each entity $E_a$ stores references to its most 'similar' entities $E_b$. For an entity with surface names $s_k$ we select its similar entities using a (non-symmetric) \textit{surface-similarity}, which we define as 
\begin{equation}
  L = \max_m \sum_k \sum_{w\in s_k\cap s_m} l(w)
\end{equation} 
where $s_m$ are the surface names of another KB entity, w is any word in $s_k$ also existing in $s_m$, and $l(w)$ is the number of characters in the word $w$. 
For each KB entity $E_a$ we find $N_S=10$ entities $E_b$ having the highest $L>0$ (may be less than 10 because of the requirement $L>0$). References to these 'surface-similar' entities $E_b$ are stored in the entity $E_a$.

In order to adjust a KB entity $E_a$, we impose a requirement on each embedding $e$ from each similar entity $E_b$: the embedding $e$ should not be able to link to the entity $E_a$ (because $E_a$ and $E_b$ are different entities).

KB is adjusted to satisfy this requirement, by iterating through all KB entities $E_a$; for each entity $E_a$ iterating through its similar entities $E_b$; for each pair $E_a$ and $E_b$ iterating through the embeddings $e_i$ of $E_a$ and $e_j$ of $E_b$. The threshold $t_i$ for $e_i$ is adjusted as:
\begin{equation}
  t_i \rightarrow c (e_j*e_i)  \;\; \text{if} \;  (e_j*e_i) > \max(T, t_i)
\end{equation} 
Here $T$ is the threshold of the entity $E_a$. We use $c = 1.01$, just enough to make the linking impossible. For clarity, the adjustment procedure is also explained by pseudo-code in Figure \ref{fig:Algo_adjustment}.    
 \begin{figure}
     \centering
     \begin{tabular}{l}
     \toprule
     {\bf{Given}}: Knowledge Base $KB = \{E_a\}$;\\ 
     Each entity $E_a$ includes:\\
     \hspace{.4cm} Threshold T = $E_a.T$\\
     \hspace{.4cm} Embeddings $E_a.embeddings = \{e_i\}$\\
     \hspace{.4cm}\hspace{.4cm} Each $e_i$ has its own threshold $t_i=-1$\\
     \hspace{.4cm} Similar entities $E_a.similar = \{E_b\}$\\
     Adjustment coefficient $c > 0$ (e.g. $c = 1.01$):\\
     Adjustment procedure:\\
     \bf{for} $E_a$ in $KB$:\\
     \hspace{.4cm}{\bf{for}} $E_b$ {\bf{in}} in $E_a.similar$:\\
     \hspace{.4cm}\hspace{.4cm}{\bf{for}} $e_i$ in $E_a.embeddings$:\\
     \hspace{.4cm}\hspace{.4cm}\hspace{.4cm}{\bf{for}} $e_j$ in $E_b.embeddings$:\\
     \hspace{.4cm}\hspace{.4cm}\hspace{.4cm}\hspace{.4cm} {\bf{if}} $(e_i*e_j) > max(E_a.T, t_i)$:\\
     \hspace{.4cm}\hspace{.4cm}\hspace{.4cm}\hspace{.4cm}\hspace{.4cm} $t_i = c * (e_i*e_j)$\\
     \hspace{.4cm}\hspace{.4cm}\hspace{.4cm}\hspace{.4cm} {\bf{if}} $(e_i*e_j) > max(E_b.T, t_j)$:\\
     \hspace{.4cm}\hspace{.4cm}\hspace{.4cm}\hspace{.4cm}\hspace{.4cm} $t_j = c * (e_i*e_j)$\\
     \bottomrule
     \end{tabular}
     \caption{Adjustment of Knowledge Base.}
     \label{fig:Algo_adjustment}
 \end{figure}
 For definiteness, we iterate through entities $E_a$ in KB in order from larger to smaller \textit{dissimilarity} of the entity; where we defined 'dissimilarity' of $E_a$ as the sum
 \begin{equation}
  \sum_{i,k}(1 - (e_i * e_k))
\end{equation}
with summation over all the pairs of embeddings from $E_a$. The motivation for this order is to start with entities that might be more prone to the conflicts and adjustments in KB. However, in the experiments described in this paper there were no any noticeable difference between this version and versions with somewhat different definitions of 'dissimilarity', and even with the opposite order of the iteration.

We do not consider here an alternative possibility: instead of increasing an embedding's threshold, we can rotate the embeddings from each other with the purpose of decreasing their product below the threshold; for more detail see Appendix \ref{sec:Appendix_AdjustByRotations}.

\section{Evaluation on Namesakes}
\label{sec:Evaluation_on_Namesakes}
Namesakes dataset consists of three parts \cite{namesakes2021}: 
\begin{enumerate}[topsep=0pt,itemsep=-1ex,partopsep=1ex,parsep=1ex]
    \item Entities: human-labeled mentions of named entities from Wikipedia entries.
    \item News: human-labeled mentions of named entities from news.
    \item Backlinks: mentions of entities linked to the entries used in Entities.
\end{enumerate}
According to \cite{Vasilyev2021Namesakes}, the mentions in all the parts are selected with the goal of creating high ambiguity of their surface names.

We are creating KB from Entities, and using News and Backlinks as sources of the mentions for evaluating NEL. These evaluation mentions are a mix of familiar and stranger mentions. The stranger mentions appear for two reasons: First, a part of the labeled mentions in News have the same surface names as the labeled mentions in Entities, but represent entities not existing in Entities.
Second, the requirement to have a certain minimal number of mentions for creating a KB entity can leave some mentions in both News and Backlinks without their counterpart KB entities. 

Performance of NEL evaluation can be represented by three indicators:
\begin{enumerate}[topsep=0pt,itemsep=-1ex,partopsep=1ex,parsep=1ex]
    \item Fraction $F_{FW}$ of familiar mentions linked to incorrect KB entity.
    \item Fraction $F_{FN}$ of familiar mentions not linked to KB.
    \item Fraction $F_{SL}$ of stranger mentions linked to KB.
\end{enumerate}
For clarity: if NEL of $N_F$ familiar mentions resulted in linking $N_{FW}$ mentions to wrong KB entities, and in not linking $N_{FN}$ mentions to KB, then $F_{FW} = N_{FW} / N_F$ and $F_{FN} = N_{FN} / N_F$. And if NEL of $N_S$ stranger mentions resulted in $N_{SL}$ linked mentions, then $F_{SL} = N_{SL} / N_S$. 

The lower each of these indicators, the better. The first indicator accounts for the worst kind of error: the mention is familiar, but it is wrongly identified. In a scenario of growing KB this would also lead to degrading KB quality.
The second indicator accounts for the most innocent error: the mention is not identified (despite it could be), but at least no wrong identity is given.
The third indicator accounts for the errors that are as bad as the first kind, with an arguable excuse that the stranger mentions are more difficult for NEL.

\section{Experiments}
\label{sec:Experiments}
\subsection{Linking to Entities of Namesakes}
\label{sec:KB_Entities}
\begin{figure}[th]
\includegraphics[width=0.48\textwidth]{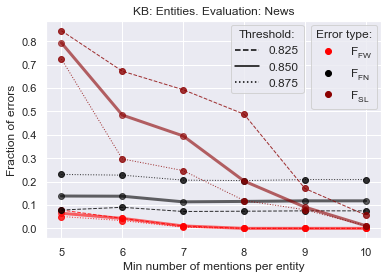}
\caption{Dependency of NEL errors on min number of mentions. KB is made of Namesakes Entities; evaluation mentions are from Namesakes News. Max number of embeddings per entity $N_E=4$. The errors are shown for threshold values $T_L = 0.825, 0.850, 0.875$.}
\label{fig:NEL_mentions_KB_Entities_News}
\end{figure}

We present here evaluation results in terms of the three indicators introduced in the previous section. 
In Figure \ref{fig:NEL_mentions_KB_Entities_News} we show the level of NEL errors for evaluating all the mentions from News.
We observe that the minimal number of mentions allowed to create a KB entity plays an important role in reducing the amount of errors, even though the entity mentions were clustered into only 4 embeddings.

The role of the linking threshold $T_L$, as expected, is the trade-off between the performance for familiar mentions and for stranger mentions. A higher threshold decreases the fraction of stranger mentions linked to KB, while increasing the fraction of familiar mentions not linked to KB. 

\begin{figure}[th]
\includegraphics[width=0.48\textwidth]{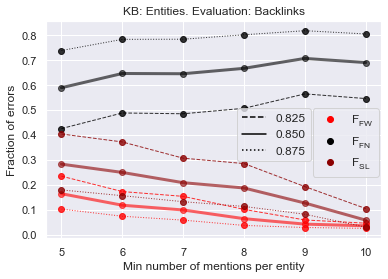}
\caption{Dependency of NEL errors on min number of mentions. KB is made of Namesakes Entities; evaluation mentions are from Namesakes Backlinks. Max number of embeddings per entity $N_E=4$. The errors are shown for threshold values $T_L = 0.825, 0.850, 0.875$.}
\label{fig:NEL_mentions_KB_Entities_Backlinks}
\end{figure}

In Figure \ref{fig:NEL_mentions_KB_Entities_Backlinks} we show again the dependency of NEL errors on the minimal number of mentions per KB entity, but now we evaluate linking of Backlinks mentions to KB (this is the only difference between the settings for the figures \ref{fig:NEL_mentions_KB_Entities_Backlinks} and \ref{fig:NEL_mentions_KB_Entities_News}).
We observe a comparable level of errors in linking a familiar mention to wrong KB entity and in wrongly linking a stranger mention to KB, but there is a much higher fraction of unlinked familiar mentions. We speculate that the reason is in the less context usually given for a named entity mention in a Wikipedia backlink, as opposed to a mention in the news.

The number of samples participating in the evaluation is presented in left panes of Tables \ref{tab:NEL_PollAdg_KB_Entities_News} and \ref{tab:NEL_PollAdg_KB_Entities_Backlinks}. Relaxed requirement on minimal number of mentions per entity allows for larger KB, and makes more familiar mentions. Backlinks part of Namesakes provides more mentions for evaluation.

\begin{figure}[th]
\includegraphics[width=0.48\textwidth]{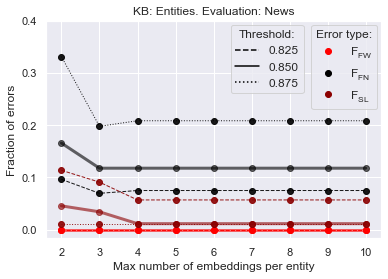}
\caption{Dependency of NEL errors on max number of embeddings. KB is made of Namesakes Entities; evaluation mentions are from Namesakes News. Min number of mentions per entity: 10. The errors are shown for threshold values $T_L = 0.825, 0.850, 0.875$.}
\label{fig:NEL_EmbLowRange_News}
\end{figure}
\begin{figure}[th]
\includegraphics[width=0.48\textwidth]{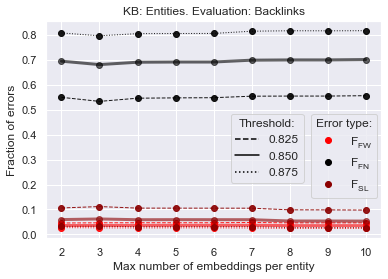}
\caption{Dependency of NEL errors on max number of embeddings. KB is made of Namesakes Entities; evaluation mentions are from Namesakes Backlinks. Min number of mentions per entity: 10. The errors are shown for threshold values $T_L = 0.825, 0.850, 0.875$.}
\label{fig:NEL_EmbLowRange_Backlinks}
\end{figure}
In Figure \ref{fig:NEL_EmbLowRange_News} we show that the limit on the number of stored embeddings can be as low as 4 - at least judging by evaluation on Namesakes News. Evaluation on Namesakes Backlinks - in Figure \ref{fig:NEL_EmbLowRange_Backlinks} - show only a very weak dependency on the maximal number of embeddings. We suggest that it may be helpful to store more embeddings, depending on the type and cleanness of the data involved in creating KB. We observe more evidence for this in Section \ref{sec:Pollution_Adjustment}. 

The number of samples involved in the evaluations in Figures \ref{fig:NEL_EmbLowRange_News} and \ref{fig:NEL_EmbLowRange_Backlinks} are presented by the last rows in the left panes in Tables \ref{tab:NEL_PollAdg_KB_Entities_News} and \ref{tab:NEL_PollAdg_KB_Entities_Backlinks} correspondingly.

\begin{figure}[th]
\includegraphics[width=0.48\textwidth]{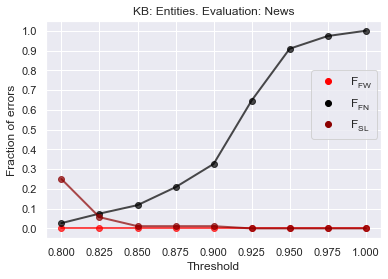}
\caption{Dependency of NEL errors on the linking threshold $T_L$. Evaluation mentions are from News. KB entities are created with min number of mentions 10, and with max number of embeddings $N_E=4$.}
\label{fig:NEL_threshold_KB_Entities_News}
\end{figure}
\begin{figure}[th]
\includegraphics[width=0.48\textwidth]{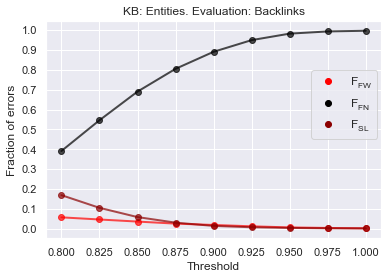}
\caption{Dependency of NEL errors on the linking threshold $T_L$. Evaluation mentions are from Backlinks. KB entities are created with min number of mentions 10, and with max number of embeddings $N_E=4$.}
\label{fig:NEL_threshold_KB_Entities_Backlinks}
\end{figure}

Increase of the linking threshold $T_L$, as expected, suppresses wrong linking, but increases the fraction of unlinked mentions. We show an example of such dependency for wide range of threshold values in Figure \ref{fig:NEL_threshold_KB_Entities_News} for NEL applied to the mentions from News, and in Figure \ref{fig:NEL_threshold_KB_Entities_Backlinks} for NEL applied to the mentions from Backlinks. The main difference between the figures is in higher fraction of unlinked familiar mentions from Backlinks; this again suggests that too many mentions in Namesakes Backlinks must have a limited context. The choice of $T_L$ should be guided by a required trade-off between the errors $F_{SL}$ vs $F_{FW}$ and $F_{FN}$.

\subsection{Linking to Polluted KB}
\label{sec:Pollution_Adjustment}

We consider here the effect of lowering KB quality on NEL. When creating KB entities, we add erroneous mentions, imitating the real life situation of not fully reliable sources. Such \textit{polluted} KB should increase NEL errors. We also expect that KB adjustment described in Section \ref{sec:KB_Adjustment} can alleviate the effect of the pollution. 

We use for pollution the "Other"-tagged mentions from Namesakes Entities \cite{Vasilyev2021Namesakes}: these mentions have the surface names of the considered entity (Wikipedia entry) but represent some different entity mentioned in the same entry. 
The pollution can be characterized by the fraction of the polluted KB entities, and by the average fraction of the wrong mentions used in creating a polluted KB entity. We set both these pollution levels to 0.5, meaning that we pollute each second entity, and that we are creating each polluted entity by making it with up to 50\% added wrong mentions (subject to availability of 'Other' mentions in the corresponding document in Namesakes Entities).

\begin{figure}[th]
\includegraphics[width=0.48\textwidth]{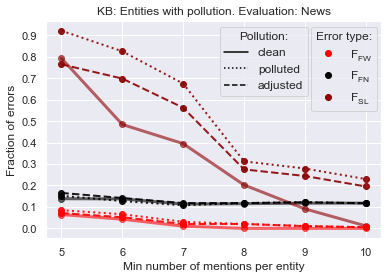}
\caption{NEL errors for linking mentions from News to KB original (solid line), polluted (dotted) and adjusted after pollution (dashed). KB entities are created with max number of embeddings $N_E=4$. Linking threshold $T_L = 0.85$.}
\label{fig:NEL_PollAdg_KB_Entities_News}
\end{figure}

In Figure \ref{fig:NEL_PollAdg_KB_Entities_News} we show how much KB pollution affects NEL, and how much our KB adjustment, described in Section \ref{sec:KB_Adjustment}, can alleviate the effect of pollution.
The pollution somewhat increases the fraction of familiar mentions linked to wrong KB entities. Pollution increases even more the fraction of stranger mentions wrongly linked to KB. The KB adjustment reduces this effect.

\begin{figure}[th]
\includegraphics[width=0.48\textwidth]{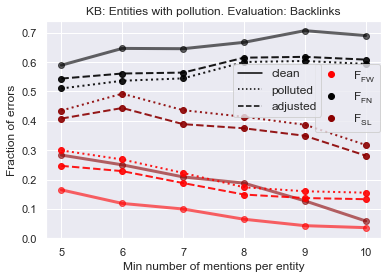}
\caption{NEL errors for linking mentions from Backlinks to KB original (solid line), polluted (dotted) and adjusted after pollution (dashed). KB entities are created with max number of embeddings $N_E=4$. Linking threshold $T_L = 0.85$.}
\label{fig:NEL_PollAdg_KB_Entities_Backlinks}
\end{figure}
Figure \ref{fig:NEL_PollAdg_KB_Entities_Backlinks} shows the evaluation for linking Backlinks mentions to the clean, polluted and polluted-and-adjusted KB.
Here we observe that the effect of pollution and adjustment on unlinked familiar mentions $F_{SL}$ is opposite: polluted KB entities encourage linking, wrong or not. Of course the effects on wrongly linked familiar mentions $F_{FW}$ and wrongly linked stranger mentions $F_{SL}$ are more important.

The number of samples participating in the evaluation is presented in Tables \ref{tab:NEL_PollAdg_KB_Entities_News} and \ref{tab:NEL_PollAdg_KB_Entities_Backlinks}.

\begin{table}[h]
\centering
\scalebox{0.75}{
\begin{tabular}{r|rrr|rrr}
 & \multicolumn{3}{c|}{\textbf{clean}} &
 \multicolumn{3}{c}{\textbf{polluted}} \\ 
 
 \textbf{M} & \textbf{KB} & \textbf{familiar} & \textbf{stranger} & \textbf{KB} & \textbf{familiar} & \textbf{stranger} \\ 
\textbf{5} & 2129 & 217 & 58 & 2440 & 224 & 51 \\
\textbf{6} & 1560 & 211 & 64 & 2034 & 212 & 63 \\
\textbf{7} & 1086 & 194 & 81 & 1509 & 195 & 80 \\
\textbf{8} & 733 & 191 & 84 & 1037 & 195 & 80 \\
\textbf{9} & 483 & 187 & 88 & 778 & 189 & 86 \\
\textbf{10} & 318 & 187 & 88 & 547 & 188 & 87 \\
\end{tabular}}
\caption{Size of KB (number of entities) and number of familiar and stranger mentions from Namesakes News participating in NEL evaluation. Here \textbf{M} is the min number of mentions allowed for creating KB entity. The left pane is for clean KB, it corresponds to Figure \ref{fig:NEL_mentions_KB_Entities_News} and to the solid line in Figure \ref{fig:NEL_PollAdg_KB_Entities_News}. The right pane is for polluted (adjusted or not) KB, it corresponds to the dotted and dashed lines in Figure \ref{fig:NEL_PollAdg_KB_Entities_News}. Total number of evaluation mentions (familiar and stranger) in each row and each pane is 275.}
\label{tab:NEL_PollAdg_KB_Entities_News}
\end{table}

\begin{table}[h]
\centering
\scalebox{0.75}{
\begin{tabular}{r|rrr|rrr}
 & \multicolumn{3}{c|}{\textbf{clean}} &
 \multicolumn{3}{c}{\textbf{polluted}} \\ 
 
 \textbf{M} & \textbf{KB} & \textbf{familiar} & \textbf{stranger} & \textbf{KB} & \textbf{familiar} & \textbf{stranger} \\ 
\textbf{5} & 2129 & 18203 & 10409 & 2440 & 22473 & 6139 \\
\textbf{6} & 1560 & 15071 & 13541 & 2034 & 20793 & 7819 \\
\textbf{7} & 1086 & 10756 & 17856 & 1509 & 15806 & 12806 \\
\textbf{8} & 733 & 8171 & 20441 & 1037 & 13345 & 15267 \\
\textbf{9} & 483 & 5528 & 23084 & 778 & 12139 & 16473 \\
\textbf{10} & 318 & 3986 & 24626 & 547 & 8895 & 19717 \\
\end{tabular}}
\caption{Size of KB (number of entities) and number of familiar and stranger mentions from Namesakes Backlinks participating in NEL evaluation. Here \textbf{M} is the min number of mentions allowed for creating KB entity. The left pane is for clean KB, it corresponds to Figure \ref{fig:NEL_mentions_KB_Entities_Backlinks} and to the solid line in Figure \ref{fig:NEL_PollAdg_KB_Entities_Backlinks}. The right pane is for polluted (adjusted or not) KB, it corresponds to the dotted and dashed lines in Figure \ref{fig:NEL_PollAdg_KB_Entities_Backlinks}. Total number of evaluation mentions (familiar and stranger) in each row and each pane is 28612.}
\label{tab:NEL_PollAdg_KB_Entities_Backlinks}
\end{table}

\begin{figure}[th]
\includegraphics[width=0.48\textwidth]{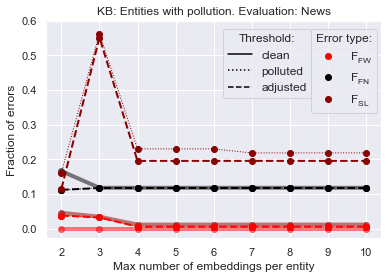}
\caption{NEL errors for linking mentions from News to KB original (solid line), polluted (dotted) and adjusted after pollution (dashed). KB entities are created with min number of mentions 10. Linking threshold $T_L = 0.85$.}
\label{fig:NEL_PollAdg_nEmbs_News}
\end{figure}

Pollution changes dependency of NEL performance on max number of entity embeddings $N_E$: in Figure \ref{fig:NEL_PollAdg_nEmbs_News} the stranger mentions errors still decrease by $N_E=7$.  
\begin{figure}[th]
\includegraphics[width=0.48\textwidth]{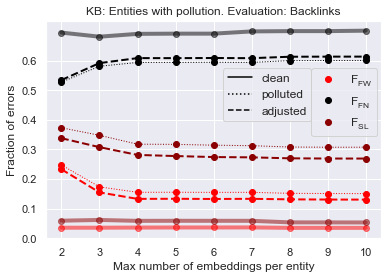}
\caption{NEL errors for linking mentions from Backlinks to KB original (solid line), polluted (dotted) and adjusted after pollution (dashed). KB entities are created with min number of mentions 10. Linking threshold $T_L = 0.85$.}
\label{fig:NEL_PollAdg_nEmbs_Backlinks}
\end{figure}
Figure \ref{fig:NEL_PollAdg_nEmbs_Backlinks} show even more change for Backlinks, as compared to almost no dependency in Figure \ref{fig:NEL_EmbLowRange_Backlinks}.

\subsection{Embeddings Tuned on More Uniform Texts}
\label{sec:ModelXSumCNNDM}
In previous subsections we observed NEL where the evaluation mentions are quite different from the mentions used in creating KB entities: KB entities were created from mentions of the entity on its own Wikipedia page; Backlinks mentions come from mentioned entities on Wikipedia backlinks; News mentions come from generic news texts. The difference between KB entities and evaluation mentions added to the difficulty of NEL.

In this subsection we consider one more variation that may cause additional difficulty for NEL: we will use a model tuned not on generic news (see Section \ref{sec:KB_Entity}), but on texts of more uniform style - a mix of texts from XSum \cite{narayan-etal-2018-dont} and CNN / Daily Mail \cite{Hermann:2015:Teaching, cnndailymail} datasets (we take the texts of the documents, not the summaries.) The details of the data and tuning are in Appendix \ref{sec:Appendix_TuningLM}.

From Figure \ref{fig:NEL_XSumCNNDM_mentions_KB_Entities_News} we observe that  the optimal value of threshold is now different, and that NEL is more sensitive to threshold.
\begin{figure}[th]
\includegraphics[width=0.48\textwidth]{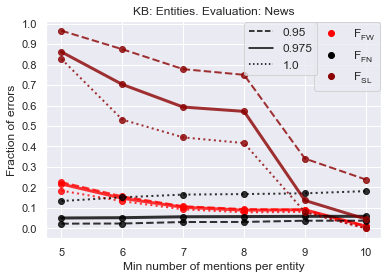}
\caption{Dependency of NEL errors on min number of mentions. Embeddings are from model tuned on XSum+CNN+DM. KB is made of Namesakes Entities; evaluation mentions are from Namesakes News. Max number of embeddings per entity $N_E=4$. The errors are shown for threshold values $T_L = 0.950, 0.975, 1.000$.}
\label{fig:NEL_XSumCNNDM_mentions_KB_Entities_News}
\end{figure}

\begin{figure}[th]
\includegraphics[width=0.48\textwidth]{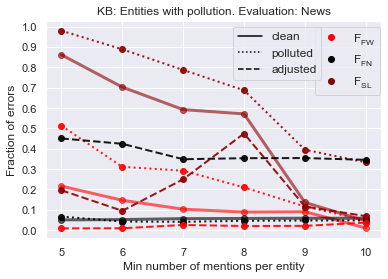}
\caption{NEL errors for linking mentions from News to KB original (solid line), polluted (dotted) and adjusted after pollution (dashed). Embeddings are from model tuned on XSum+CNN+DM. KB entities are created with max number of embeddings $N_E=4$. Linking threshold $T_L = 0.975$.}
\label{fig:NEL_XSumCNNDM_PollAdg_KB_Entities_News}
\end{figure}

From Figure \ref{fig:NEL_XSumCNNDM_PollAdg_KB_Entities_News} we observe that effect of KB adjustment can be stronger. The adjustment, while strongly reducing the most serious errors - $F_{FW}$ and $F_{SL}$ - increases the fraction of unlinked familiar mentions $F_{FN}$. The latter is understandable as a conseqeunce of increasing individual embedding thresholds in KB entities.

\begin{figure}[th]
\includegraphics[width=0.48\textwidth]{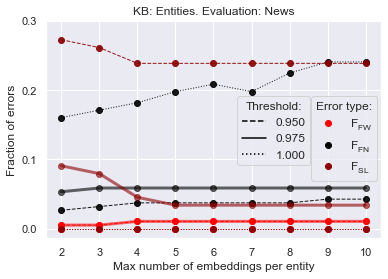}
\caption{NEL errors for linking mentions from News to KB. Embeddings are from model tuned on XSum+CNN+DM. KB entities are created with min number of mentions 10. The errors are shown for linking threshold values $T_L = 0.950, 0.975, 1.000$.}
\label{fig:NEL_XSumCNNDM_Embs_3thrs_News}
\end{figure}

From Figure \ref{fig:NEL_XSumCNNDM_Embs_3thrs_News} we observe that choosing the KB entity's max number of embeddings $N_E$ may also be affected by the type of model used for embeddings (compare with Figure \ref{fig:NEL_EmbLowRange_News}).

\section{Conclusion}
We introduced a simple and practical representation of named entities by multiple embeddings. We reviewed NEL performance for linking text mentions to KB of such entities, using Namesakes dataset \cite{Vasilyev2021Namesakes} as the source both for the mentions and for building the KB. As a dataset of ambiguous named entities, Namesakes makes NEL difficult and helps to reveal the errors and to observe the behavior and the dependencies of our NEL method.

We observed that a requirement of a minimal number of mentions for creating KB entity is important for NEL performance: a requirement of minimum 10 mentions gives much better results than more relaxed settings (Figures \ref{fig:NEL_mentions_KB_Entities_News}, \ref{fig:NEL_mentions_KB_Entities_Backlinks}).
We described a KB adjustment based on only KB data; we have shown that the adjustment helps to reduce NEL errors when KB entities are polluted by admix of wrong mentions (Figures \ref{fig:NEL_PollAdg_KB_Entities_News}, \ref{fig:NEL_PollAdg_KB_Entities_Backlinks}, \ref{fig:NEL_XSumCNNDM_PollAdg_KB_Entities_News}).

Through the paper we evaluated our NEL method with static small KB (albeit on intentionally challenging data). More general scenarios can be considered, such as growing and adjusting KB with linked mentions. Even within the limits of static KB, there are interesting issues left out of our consideration here. The KB pollution can be moderated by heuristic algorithms that filter the mentions for KB entities. Also, in realistic ingestion of data to large scale KB (e.g. all Wikipedia named entities), there are much more available mentions for some entities. From our preliminary observations (not included in the paper) we speculate that the optimal number of embeddings can be higher but still within reasonable limits (up to 20), and that the effect of KB adjustment can be stronger.

\section*{Limitations}
We acknowledge the following limitations of this work:
\begin{enumerate}[topsep=0pt,itemsep=-1ex,partopsep=1ex,parsep=1ex]
    \item Our observations are limited to the case of fixed KB. We left out the consideration of accepting new linked mentions to KB and of the corresponding changes in KB entity representations, and in NEL quality.  
    \item We considered linking of 'lone' mentions, not using advantages of coreference with mentions of the same entity or related entities in the text. 
    \item We investigated the behavior and dependencies of NEL with our representations on a specific dataset (Namesakes), the dataset is being challenging by having high concentration of namesakes. Evaluation of the method on a large scale KB (for example, KB of all Wikipedia named entities or of entities from some other documents) is left out of this paper. An immediate difference is that such KB is inevitably polluted, albeit not as much as in our considerations in the paper.
    \item We explored linking of mentions from two kinds of texts: news and backlinks, the latter is fairly artificial example. Realistically, mentions come from documents of very different formats and styles. While we did imitated here the difference between the mentions on which KB is built (the mentions from their own Wikipedia entries) and the evaluation mentions (the mentions from news or backlinks), this is still just a two examples.
\end{enumerate}

\section*{Acknowledgments}
We thank Randy Sawaya for review of the paper and valuable feedback.

\bibliography{anthology,custom}
\bibliographystyle{acl_natbib}

\appendix
\section{Tuning LM on Named Entities}
\label{sec:Appendix_TuningLM}
\subsection{Tuning}
\label{ssec:Tuning}
We obtained embeddings by using a pretrained language model (LM) tuned on located mentions of entities (Section \ref{sec:KB_Entity}). The texts for the tuning (9831 texts) were taken randomly from generic news, and processed by the named entity recognition (NER) model 'dbmdz/bert-large-cased-finetuned-conll03-english', accessed via the transformers library \citep{Wolf2020Transformers}). (As explained in Introduction, we are focused on NEL under the assumption that the mentions of named entities are already located in the text.)

The purpose is only in tuning the pretrained model for doing LM task on already located mentions of named entities, without specializing on some specific dataset of named entities. The number of different identified types of the located mentions were comparable: approximately 7.6 locations, 10.4 persons, 11.8 organizations and 6.4 miscellaneous named entities per text, with the average length of text 3300 characters. In this work we do not use the types of locations.   

The located mentions in the texts are marked by enveloping them into square brackets, e.g. "... Minority Leader [Kevin McCarthy] and other..." (after making sure in advance that any such brackets in the text are replaced by round brackets). This procedure is done both for tuning LM, and for inference. During the tuning the name in the brackets is sometimes (with probability 0.5) replaced by another random name from the text. At inference there are no replacements (Section \ref{ssec:Inference}). 

The labels for tuning are all the tokens within all the square brackets that happen to occur (located by NER model) within the input of LM. We tuned the pretrained 'bert-base-uncased' LM \citep{Wolf2020Transformers} on texts from random daily news. Each input for tuning is composed of whole sentences - as many sentences as fits into the maximal input size (512 tokens for the LM).

\subsection{Inference}
\label{ssec:Inference}
At inference the procedure is similar: The text must be processed by NER model; the recognized named entities must be bracketed '[]' (but never replaced). Then the text must be processed by chunks, each chunk consisting of whole sentences - as many sentences as fits into the LM maximal input size. Then for each named entity mention (name in the brackets) the embedding is taken for the first token of the mention, from the last hidden layer of LM.

However, for the inference in our experiments here on Namesakes, we are using already recognized and labeled named entities of Namesakes, so the step with running NER is not needed. We have all the mentions already located, both for creating KB entities and for NEL evaluation.

\subsection{Tuning on texts from XSum, CNN / Daily Mail}
\label{ssec:Tuning_XSumCNNDM}
In order to have a model trained on less varied, more uniform style texts, we also tuned the same model bert-base-uncased on a mix of texts from well known datasets XSum \cite{narayan-etal-2018-dont} and CNN / Daily Mail \cite{Hermann:2015:Teaching, cnndailymail}. (We observed effects of switching to such model in Section \ref{sec:ModelXSumCNNDM}.)

We created a training set of 33,000 texts comprised of 11,000 randomly selected texts from each of the primary sources (XSum, CNN, and Daily Mail) with a validation set of 1,500 texts as well as a test set of 1,500 texts (500 texts randomly selected from each primary source). Entity extraction and model tuning were conducted using the same models and strategy described in \ref{ssec:Tuning}. This resulted in a training set of 65,530 individual samples (a single sample being a context window of 512 tokens containing at least one entity) with a validation set of 3,005 samples and a test set of 3,030 samples. Model tuning was conducted for a single epoch on an NVIDIA Tesla T4 GPU, and optimized for the validation set.

\section{Clustering Embeddings for KB Entity}
\label{sec:Appendix_Settings}
As explained in Section \ref{sec:KB_Entity}, KB entity stores a limited number $N_E$ of embeddings. When the number of available reliable mentions exceeds $N_E$, the corresponding normalized embeddings are clustered. For this purpose we are using agglomerative clustering with euclidean affinity and with average linkage.

From each cluster we select one representation embedding: the embedding closest (by euclidean distance) to the center of the cluster. The center is defined as the average of all embeddings of the cluster.

\section{Adjustment of KB by rotation of embeddings}
\label{sec:Appendix_AdjustByRotations}
KB adjustment considered in the paper is defined in Section \ref{sec:KB_Adjustment}. We adjusted individual thresholds for entity representation embeddings, in order to prevent linking of one KB entity to another. Here we point out an alternative possibility to prevent such linking: we can rotate embeddings away from each other.

Suppose that an embedding $e_j$ from an entity $E_b$ is able to link to the entity $E_a$, by being too similar:
\begin{equation}
  (e_j*e_i) > T
\end{equation}
Here $e_i$ is one of embeddings of $E_a$, and $T$ is the threshold of $E_a$. We can then replace the embedding $e_i$ by the embedding
\begin{equation}
  e_i^* = cos(\alpha)e_i + sin(\alpha)e_j
\end{equation}
with the goal of having
\begin{equation}
  s \equiv (e_j^**e_i) = T / c \equiv t; \hspace{20pt} c > 1
\end{equation}
where the coefficient $c$ is slightly higher than $1$, for example $c = 1.01$. The solution is simple:
\begin{equation}
  cos(\alpha) = \frac{ts + \sqrt{1 + s^2 - t^2}}{1 + s^2}
\end{equation}
\begin{equation}
  sin(\alpha) = - \sqrt{1 - cos(\alpha)}
\end{equation}
There can be also a variation where we change both $e_j$ and $e_i$. Such alternatives are as simple for processing as the adjustment version that we have considered through the paper. However, the adjustment of thresholds can be undone and redone at any time, while to do the same after rotating embeddings we would have to store the original embeddings. This makes the rotation version more expensive for storage. Performance-wise, for NEL on Namesakes, we have not seen advantages of rotations over threshold adjustments.

Both the thresholds adjustment version and the adjustment by rotations version can be made more 'iterative': Each individual adjustment (the increase of a threshold or the angle of rotation) can be made smaller, but there would be multiple passes of iterations over all KB entities - possibly even in random order. The passes stop when the number of adjustments made over the pass is zero or below a certain limit. In case of larger data such versions may give better results, but the expensiveness of multiple passes makes this impractical.

\end{document}